\documentclass{article}
\usepackage{cmap}\usepackage[T1]{fontenc} 
\usepackage[table,x11names]{xcolor} 
\usepackage{iclr2016_conference}
\usepackage[utf8]{inputenc}
\usepackage{amsmath}
\usepackage{amssymb}
\usepackage{amsthm}
\usepackage{mathrsfs}
\usepackage{url}
\usepackage{tabularx}
\usepackage[pagebackref,breaklinks=true]{hyperref} 
\usepackage{caption}    
\renewcommand*\backref[1]{\ifx#1\relax \else (\^ #1) \fi} 
\usepackage{breakcites}
\usepackage{graphicx}   
\graphicspath{ {img/} }
\usepackage{sidecap}
\usepackage{changepage} 
\usepackage{pifont}     
\usepackage{multirow}   
\usepackage{array}      
\newcolumntype{C}[1]{>{\centering\arraybackslash}m{#1}}
\newcolumntype{Y}{>{\centering\arraybackslash}X} 
\let\vec\mathbf 
\usepackage{marvosym} 

\newcommand{\R}{\mathbb{R}} 
\newcommand{\E}{\mathbb{E}} 
\renewcommand{\L}{\mathcal{L}} 
\newcommand{\dif}{\,\mathrm{d}} 
\DeclareMathOperator*{\argmin}{arg\,min}
\newcommand{\crss}{\ding{54}} 

\theoremstyle{definition}
\newtheorem{definition}{Definition}

\newcommand{\cData}[1]{\textcolor[rgb]{.65,.05,.05}{#1}}
\newcommand{\cArch}[1]{\textcolor[rgb]{0.75,0.5,0.05}{#1}}
\newcommand{\cError}[1]{\textcolor[rgb]{0,0.65,0.15}{#1}}
\newcommand{\cReg}[1]{\textcolor[rgb]{0.05,0.45,0.75}{#1}}
\newcommand{\cOpt}[1]{\textcolor[rgb]{0.5,0.2,0.9}{#1}}

\title{Regularization for Deep Learning:\\A~Taxonomy}
\author{Jan Kukačka, Vladimir Golkov, and Daniel Cremers \\
\texttt{\{jan.kukacka, vladimir.golkov, cremers\}@tum.de} \\
Computer Vision Group \\
Department of Informatics \\
Technical University of Munich}
\date{June 2017}

\begin{document}

\maketitle

\begin{abstract}
Regularization is one of the crucial ingredients of deep learning, yet the term \emph{regularization} has various definitions, and regularization methods are often studied separately from each other. In our work we present a systematic, unifying taxonomy to categorize existing methods. We distinguish methods that affect data, network architectures, error terms, regularization terms, and optimization procedures. We do not provide all details about the listed methods; instead, we present an overview of how the methods can be sorted into meaningful categories and sub-categories. This helps revealing links and fundamental similarities between them. Finally, we include practical recommendations both for users and for developers of new regularization methods.
\end{abstract}

\section{Introduction}

Regularization is one of the key elements of machine learning, particularly of deep learning~\citep{dlb}, 
allowing to generalize well to unseen data even when training on a finite training set or with an imperfect optimization procedure.
In the traditional sense of optimization and also in older neural networks literature, the term ``regularization'' is reserved solely for a penalty term in the loss function~\citep{bishop95b}. Recently, the term has adopted a broader meaning:~\citet[Chap.~5]{dlb} loosely define it as \textit{``any modification we make to a learning algorithm that is intended to reduce its test error but not its training error''}. We find this definition slightly restrictive and present our working definition of regularization, since many techniques considered as regularization do reduce the training error (e.g.~weight decay in AlexNet~\citep{krizhevsky12}).
\begin{definition}
\label{def:reg}
\textbf{Regularization}
is any supplementary technique that aims at making the model generalize better, i.e.~produce better results on the test set.
\end{definition}

This can include various properties of the loss function, the loss optimization algorithm, or other techniques. Note that this definition is more in line with machine learning literature than with inverse problems literature, the latter using a more restrictive definition.

Before we proceed to the presentation of our taxonomy, we revisit some basic machine learning theory in Section~\ref{sec:theory}. This will provide a justification of the top level of the taxonomy. In Sections~\mbox{\ref{sec:data}--\ref{sec:opt}}, we continue with a finer division of the individual classes of the regularization techniques, followed by our practical recommendations in Section~\ref{sec:discussion}. We are aware that the many research works discussed in this taxonomy cannot be summarized in a single sentence.  For the sake of structuring the multitude of papers, we decided to merely describe a certain subset of their properties according to the focus of our taxonomy.

\section{Theoretical framework}
\label{sec:theory}
The central task of our interest is model fitting: finding a function $f$ that can well approximate a desired mapping 
from inputs $x$ to desired outputs $f(x)$. A~given input $x$ can have an associated target $t$ which dictates the desired output $f(x)$ directly (or in some applications indirectly~\citep{ulyanov16,johnson16}).
A~typical example of having available targets $t$ is supervised learning. Data samples $(x,t)$ then follow a ground truth probability distribution $P$.

In many applications, neural networks have proven to be a good family of functions to choose $f$ from. A~neural network is a function $f_w: x \mapsto y$ with trainable weights $w \in W$. \textit{Training} the network means finding a weight configuration $w^*$ minimizing a \textit{loss function} $\L:W \rightarrow \R$ as follows:
\begin{equation} 
\label{eq:wminL}
w^\ast = \argmin_w \L(w) .
\end{equation}
Usually the loss function takes the form of \textit{expected risk}:
\begin{equation} 
\label{eq:exprisk}
\begin{split}
\L  = \E_{(x,t)\sim P} \Big[ E\big(f_w(x),t\big) + R(\ldots) \Big],
\end{split}
\end{equation}
where we identify two parts, an \textit{error function} $E$ 
and a \textit{regularization term} $R$. The error function depends on the targets and assigns a penalty to model predictions according to their \textit{consistency} with the targets. The regularization term assigns a penalty to the model based on other criteria. It may depend on anything except the targets, for example on the weights (see Section~\ref{sec:regterm}).

The expected risk cannot be minimized directly since the data distribution $P$ is unknown. Instead, a \textit{training set} $\mathcal{D}$ sampled from the distribution is given. The minimization of the expected risk can be then approximated by minimizing the \textit{empirical risk}~$\hat\L$:
\begin{equation}
\label{eq:emprisk}
\cOpt{\argmin_{\textcolor[rgb]{0.0,0.0,0.0}w}} \, \frac{1}{|\mathcal{D}|} \sum_{\cData{(x_i,t_i)\in \mathcal{D}}}  \cError{E}\big(\cArch{f}_w(\cData{x_i}),\cData{t_i}\big) + \cReg{R(\ldots)}
\end{equation}
where $(x_i,t_i)$ are samples from $\mathcal{D}$.

Now we have the minimal background to formalize the division of regularization methods into a systematic taxonomy. In the minimization of the empirical risk, Eq.~\eqref{eq:emprisk}, we can identify the following elements that are responsible for the value of the learned weights, and thus can contribute to regularization:
\begin{itemize}
    \item[\cData{\textbullet}] $\mathcal{D}$: The training set, discussed in Section~\ref{sec:data}
    \item[\cArch{\textbullet}] $f$: The selected model family, discussed in Section~\ref{sec:arch}
    \item[\cError{\textbullet}] $E$: The error function, briefly discussed in Section~\ref{sec:errorterm}
    \item[\cReg{\textbullet}] $R$: The regularization term, discussed in Section~\ref{sec:regterm}
    \item[\cOpt{\textbullet}] The optimization procedure itself, discussed in Section~\ref{sec:opt}
\end{itemize}
Ambiguity regarding the splitting of methods into these categories and their subcategories is discussed in Appendix~\ref{sec:ambi} using notation from Section~\ref{sec:data}.

\section{Regularization via data}
\label{sec:data}

The quality of a trained model depends largely on the training data. Apart from acquisition/selection of appropriate training data, it is possible to employ regularization via data. This is done by applying some transformation to the training set $\mathcal{D}$, resulting in a new set $\mathcal{D}_R$. Some transformations perform feature extraction or pre-processing, modifying the feature space or the distribution of the data to some representation simplifying the learning task. Other methods allow generating new samples to create a larger, possibly infinite, \textit{augmented} dataset. These two principles are somewhat independent and may be combined. The goal of regularization via data is either one of them, or the other, or both.
They both rely on \textit{transformations with (stochastic) parameters}:
\begin{definition}
\textbf{Transformation with stochastic parameters} is a function $\tau_\theta$ with parameters $\theta$ which follow some probability distribution.
\end{definition}
In this context we consider $\tau_\theta$ which can operate on network inputs, activations in hidden layers, or targets. An example of a transformation with stochastic parameters is the corruption of inputs by Gaussian noise~\citep{bishop95,an96}:
\begin{equation}
\label{eq:noiseoninput}
\tau_\theta(x) =  x + \theta, \quad \theta \sim \mathcal{N}(\vec{0},\vec\Sigma).
\end{equation}
The stochasticity of the transformation parameters is responsible for generating new samples, i.e.~\textit{data augmentation}. Note that the term \textit{data augmentation} often refers specifically to transformations of inputs or hidden activations, but here we also list transformations of targets for completeness. The exception to the stochasticity is when $\theta$ follows a delta distribution, in which case the transformation parameters become deterministic and the dataset size is \emph{not} augmented.

We can categorize the data-based methods according to the properties of the used transformation and of the distribution of its parameters. We identify the following criteria for categorization (some of them later serve as columns in Tables~\ref{tbl:data:generic}--\ref{tbl:data:domainspecific}):

\paragraph{Stochasticity of the transformation parameters $\theta$}
\label{sec:data:stoch}
\begin{itemize}
    \item Deterministic parameters: Parameters $\theta$ follow a delta distribution, size of the dataset remains unchanged 
    \item Stochastic parameters: Allow generation of a larger, possibly infinite, dataset. Various strategies for sampling of $\theta$ exist:
    \label{sec:data:sampling}
    \begin{itemize}
        \item Random: Draw a random $\theta$ from the specified distribution
        \item Adaptive: Value of $\theta$ is the result of an optimization procedure, usually with the objective of maximizing the network error on the transformed sample (such ``challenging'' sample is considered to be the most informative one at current training stage),
        or minimizing the difference between the network prediction and a predefined fake target $t'$
        \begin{itemize}
            \item Constrained optimization: $\theta$ found by maximizing error under hard constraints (support of the distribution of $\theta$ controls the strongest allowed transformation)
            \item Unconstrained optimization: $\theta$ found by maximizing modified error function, using 
            the distribution of $\theta$ as weighting
            (proposed herein for completeness, not yet tested)
            \item Stochastic:  $\theta$ found by taking a fixed number of samples of $\theta$ and using the one yielding the highest error
        \end{itemize}    
    \end{itemize}
\end{itemize}

\paragraph{Effect on the data representation}
\label{sec:data:repr}
\begin{itemize}
    \item Representation-preserving transformations: Preserve the feature space and attempt to preserve the data distribution
    \item Representation-modifying transformations:  Map the data to a different representation (different distribution or even new feature space) that may disentangle the underlying factors of the original representation and make the learning problem easier
\end{itemize}

\paragraph{Transformation space}
\label{sec:data:space}
\begin{itemize}
    \item Input: Transformation is applied to $x$
    \item Hidden-feature space: Transformation is applied to some deep-layer representation of samples (this also uses parts of $f$ and $w$ to map the input into the hidden-feature space; such transformations act inside the network $f_w$ and thus can be considered part of the architecture, additionally fitting Section~\ref{sec:arch})
    \item Target: Transformation is applied to $t$ (can only be used during the \hyperref[sec:data:phase]{training phase} since labels are not shown to the model at test time)
\end{itemize}

\paragraph{Universality}
\label{sec:data:universality}
\begin{itemize}
    \item Generic: Applicable to all data domains
    \item Domain-specific: Specific (handcrafted) for the problem at hand, for example image rotations
\end{itemize}

\paragraph{Dependence of the distribution of $\theta$}
\label{sec:data:cond}
\begin{itemize}
    \item $p(\theta)$: distribution of $\theta$ is the same for all samples
    \item $p(\theta|t)$: distribution of $\theta$ can be different for each target (class)
    \item $p(\theta|t')$: distribution of $\theta$ depends on desired (fake) target $t'$
    \item $p(\theta|x)$: distribution of $\theta$ can be different for each input vector (with implicit dependence on~$f$ and~$w$ if the transformation is in \hyperref[sec:data:space]{hidden-feature space})
    \item $p(\theta|\mathcal{D})$: distribution of $\theta$ depends on the whole training dataset
    \item $p(\theta|\vec{x})$: distribution of $\theta$ depends on a batch of training inputs 
    (for example (parts~of) the current mini-batch, or also previous mini-batches)
    \item $p(\theta|\mathrm{time})$: distribution of $\theta$ depends on time (current training iteration)
    \item $p(\theta|\pi)$: distribution of $\theta$ depends on some trainable parameters $\pi$ subject to loss minimization (i.e.~the parameters $\pi$ evolve during training along with the network weights $w$)
    \item Combinations of the above, e.g.~$p(\theta|x,t)$, $p(\theta|x,\pi)$, $p(\theta|x,t')$, $p(\theta|x,\mathcal{D})$, $p(\theta|t,\mathcal{D})$, $p(\theta|x,t,\mathcal{D})$
\end{itemize}

\paragraph{Phase}
\label{sec:data:phase}
\begin{itemize}
    \item Training: Transformation of training samples
    \item Test: Transformation of test samples, for example multiple augmented variants of a sample are classified and the result is aggregated over them
\end{itemize}

\begin{table}
\setlength\arrayrulewidth{.1pt}
    \footnotesize{
     \begin{tabularx}{\textwidth}{@{}|>{\raggedright\arraybackslash}X|p{1.75cm}|>{\raggedright\arraybackslash}p{2.25cm}|p{1.95cm}|p{1.25cm}|@{}}
        \hline
        Method  &  \hyperref[sec:data:cond]{Dependence} & \hyperref[sec:data:space]{Transformation space} & \hyperref[sec:data:sampling]{Stochasticity\newline($\theta$ sampling)} & \hyperref[sec:data:phase]{Phase} \\
        \hline \hline
        Gaussian noise on input\newline\citep{bishop95b,an96} & 
            $p(\theta)$ & Input & Random & Training \\ \hline
        Gaussian noise on hidden units\newline\citep{devries17} &
            $p(\theta)$ & Hidden features & Random & Training \\ \hline
        Dropout \citep{hinton12, srivastava14} &
            $p(\theta)$ & Input and\newline hidden features & Random & Training \\ \hline
        Random dropout probability\newline\citep[Sec.~4]{bouthillier15} &
            $p(\theta)$ & Input and\newline hidden features & Random & Training \\ \hline
        Curriculum dropout\newline\citep{morerio17}  &
            $p(\theta|\mathrm{time})$ & Input and\newline hidden features & Random & Training \\ \hline
        Bayesian dropout\newline \citep{maeda14} &
            $p(\theta|\pi)$ & Input and\newline hidden features & Random & Training \\ \hline
        Standout (adaptive dropout)\newline\citep{ba13} & 
            $p(\theta|x,\pi)$ & Input and\newline hidden features & Random & Training \\ \hline
        ``Projection'' of dropout noise into input space \citep[Sec.~3]{bouthillier15} &
            $p(\theta|x,f,w)$
            & Input {\scriptsize\newline Uses auxiliary $\tau$ in hidden-feature space.} & Random & Training \\ \hline
        Approximation of Gaussian process by test-time dropout\newline\citep{gal16} &
            $p(\theta)$ & Input and\newline hidden features & Random & Test \\ \hline
        Stochastic depth~\citep{huang16} &
            $p(\theta)$ & Hidden features & Random & Training \\ \hline
        Noisy activation functions\newline\citep{nair10,xu15,gulcehre16b} &
            $p(\theta|x)$ & Hidden features & Random & Training \\ \hline
        Training with adversarial examples \citep{szegedy14}&
            $p(\theta|x, t')$ & Input & Adaptive\newline\tiny{Constrained} & Training \\ \hline
        Network fooling (adversarial examples) \citep{szegedy14}\newline
            \emph{(Not for regularization)} &
            $p(\theta|x, t')$ & Input & Adaptive\newline\tiny{Constrained} & Test \\ \hline
        Synthetic minority oversampling in hidden-feature space~\citep{wong16} &
            $p(\theta|x,t,\mathcal{D})$ & Hidden features & Random & Training \\ \hline
        Inter- and extrapolation in hidden-feature space~\citep{devries17} & 
            $p(\theta|x,t,\mathcal{D})$ & Hidden features  & Random & Training \\ \hline
        Batch normalization \citep{ioffe15}, Ghost batch normalization \citep{hoffer17} &
            $p(\theta|\vec{x})$ & Hidden features & De\-ter\-min\-is\-tic & Training and test \\ \hline
        Layer normalization \newline\citep{ba16} &
            $p(\theta|x)$ & Hidden features & De\-ter\-min\-is\-tic & Training and test \\ \hline
        Annealed noise on targets\newline\citep{wang99} &
            $p(\theta|\mathrm{time})$ & Target & Random & Training \\ \hline
        Label smoothing
        \citetext{\citealp[Sec.~7]{szegedy16}; \citealp[Chap.~7]{dlb}} &
            $p(\theta)$ & Target & De\-ter\-min\-is\-tic & Training \\ \hline
        Model compression (mimic models, distilled models) \citep{bucila06,ba14,hinton15} & 
            $p(\theta|x,\mathcal{D})$ & Target & De\-ter\-min\-is\-tic & Training \\ \hline
    \end{tabularx}
    } 
    \caption{\label{tbl:data:generic}
    Existing \hyperref[sec:data:universality]{\emph{generic}} data-based methods classified according to our taxonomy. Table columns are described in Section~\ref{sec:data}.} 
\end{table}

\begin{table}
\setlength\arrayrulewidth{.1pt}
    \footnotesize{
     \begin{tabularx}{\textwidth}{@{}|>{\raggedright\arraybackslash}X|p{1.75cm}|>{\raggedright\arraybackslash}p{2.25cm}|p{1.95cm}|p{1.25cm}|@{}}
        \hline
        Method  &  \hyperref[sec:data:cond]{Dependence} & \hyperref[sec:data:space]{Transformation space} & \hyperref[sec:data:sampling]{Stochasticity\newline($\theta$ sampling)} & \hyperref[sec:data:phase]{Phase} \\
        \hline \hline 
        Rigid and elastic image transformation \citep{baird90,yaegger96,simard03,ciresan10} & 
            $p(\theta)$ & Input & Random & Training \\ \hline
        Test-time image transformations \citep{simonyan15,dieleman15} &
            $p(\theta)$ & Input & Random & Test \\ \hline
        Sound transformations\newline\citep{salamon17} &
            $p(\theta)$ & Input & Random & Training \\ \hline
        Error-maximizing rigid image transformations\newline\citep{loosli07, fawzi16} &
            $p(\theta)$ & Input & Adaptive \tiny{stochastic \&\newline constrained, respectively} & Training \\ \hline
        Learning class-specific elastic image-deformation fields\newline\citep{hauberg16} &
            $p(\theta|t,\mathcal{D})$ & Input & Random & Training \\ \hline
         Any handcrafted data preprocessing, for example scale-invariant feature transform (SIFT) for images \citep{lowe99} &
            $p(\theta)$ & Input & De\-ter\-min\-is\-tic & Training and test \\ \hline
        Overfeat \citep{sermanet13} &
            $p(\theta)$ & Input & De\-ter\-min\-is\-tic & Training and test \\ \hline
    \end{tabularx}
    } 
    \caption{\label{tbl:data:domainspecific}Existing \hyperref[sec:data:universality]{\emph{domain-specific}} data-based methods classified according to our taxonomy. Table columns are described in Section~\ref{sec:data}. 
    Note that these methods are never applied on the hidden features, because domain knowledge cannot be applied on them.
    }
\end{table}

A~review of existing methods that use \hyperref[sec:data:universality]{generic transformations} can be found in Table~\ref{tbl:data:generic}. Dropout in its original form~\citep{hinton12, srivastava14} is one of the most popular methods from the generic group, but also several variants of Dropout have been proposed that provide additional theoretical motivation and improved empirical results (Standout~\citep{ba13}, Random dropout probability~\citep{bouthillier15}, Bayesian dropout~\citep{maeda14}, Test-time dropout~\citep{gal16}).

Table~\ref{tbl:data:domainspecific} contains a list of some \hyperref[sec:data:universality]{domain-specific} methods focused especially on the image domain. Here the most used method is rigid and elastic image deformation.

\paragraph{Target-preserving data augmentation} In the following, we discuss an important group of methods: \textit{target-preserving data augmentation}. These methods use \hyperref[sec:data:stoch]{\textbf{stochastic}} transformations in \hyperref[sec:data:space]{\textbf{input} and \textbf{hidden-feature} spaces}, while preserving the original target~$t$. As can be seen in the respective two columns in Tables~\ref{tbl:data:generic}--\ref{tbl:data:domainspecific}, most of the listed methods have exactly these properties. These methods transform the training set to a distribution $Q$, which is used for training instead. In other words, the training samples $(x_i,t_i)\in\mathcal{D}$ are replaced in the empirical risk loss function (Eq.~\eqref{eq:emprisk}) by augmented training samples $(\tau_\theta(x_i), t_i) \sim Q$. By randomly sampling the transformation parameters~$\theta$ and thus creating many new samples $(\tau_\theta(x_i), t_i)$ from each original training sample $(x_i,t_i)$, data augmentation attempts to bridge the limited-data gap between the expected and the empirical risk, Eqs.~\eqref{eq:exprisk}--\eqref{eq:emprisk}.
While unlimited sampling from $Q$ provides more data than the original dataset $\mathcal{D}$, both of them usually are merely approximations of the ground truth data distribution or of an ideal training dataset; both $\mathcal{D}$ and $Q$ have their own distinct biases, advantages and disadvantages. For example, elastic image deformations result in images that are not perfectly realistic; this is not necessarily a disadvantage, but it is a bias compared to the ground truth data distribution; in any case, the advantages (having more training data) often prevail. In some cases, it may be even desired for $Q$ to be deliberately \emph{different} from the ground truth data distribution. For example, in case of class imbalance (unbalanced abundance or importance of classes), a common regularization strategy is to undersample or oversample the data, sometimes leading to a less realistic $Q$ but better models. This is how an \emph{ideal training dataset} may be different from the \emph{ground truth data distribution}.

If the transformation is additionally \hyperref[sec:data:repr]{\textbf{representation-preserving}}, then the distribution $Q$ created by the transformation~$\tau_\theta$ attempts to mimic the ground truth data distribution~$P$. Otherwise, the notion of a ``ground truth data distribution'' in the modified representation may be vague. We provide more details about the transition from $\mathcal{D}$ to $Q$ in Appendix~\ref{sec:aug}.

\paragraph{Summary of data-based methods} Data-based regularization is a popular and very useful way to improve the results of deep learning. In this section we formalized this group of methods and showed that seemingly unrelated techniques such as Target-preserving data augmentation, Dropout, or Batch normalization are methodologically surprisingly close to each other. In Section~\ref{sec:discussion} we discuss future directions that we find promising.

\section{Regularization via the network architecture}
\label{sec:arch}
A~network architecture~$f$ can be selected to have certain properties or match certain assumptions in order to have a regularizing effect.\footnote{The network architecture is represented by a function $f: (w,x) \mapsto y$, and together with the set~$W$ of all its possible weight configurations defines a set of mappings that this particular architecture can realize: $\{f_w: x \mapsto y \mid \forall w \in W\}$.}

\begin{table}
\setlength\arrayrulewidth{.1pt}
    \tiny { 
    \begin{tabularx}{\textwidth}{|@{}>{\raggedright\arraybackslash}p{3.5cm}|p{1.8cm}|X@{}|}
        \hline
        Method  &  Method class & Assumptions about an appropriate learnable input-output mapping \\ \hline \hline
            Any chosen (not overly complex) architecture & * &
                Mapping can be well approximated by functions from the chosen family which are easily accessible by optimization. \\ \hline
            Small network & * &
                Mapping is simple (complexity of the mapping depends on the number of network units and layers). \\ \hline
            Deep network & * &
                The mapping is complex, but can be decomposed into a composition (or generally into a directed acyclic graph) of simple nonlinear transformations, e.g.\ affine transformation followed by simple nonlinearity (fully-connected layer), ``multi-channel convolution'' followed by simple nonlinearity (convolutional layer), etc. \\ \hline
            Hard bottleneck (layer with few neurons); soft bottleneck (e.g.\ Jacobian penalty \citep{rifai11a}, see Section~\ref{sec:regterm}) & Layer operation &
                Data concentrates around a lower-dimensional manifold; has few factors of variation. \\ \hline
            Convolutional networks \citetext{\citealp{fukushima82}; \citealp[pp.~348-352]{rumelhart86}; \citealp{lecun89}; \citealp{simard03}} & Layer operation &
                Spatially local and shift-equivariant feature extraction is all we need. \\ \hline
            Dilated convolutions\newline\citep{yu15}
                & Layer operation &
                Like convolutional networks. Additionally: Sparse sampling of wide local neighborhoods provides relevant information, and better preserves relevant high-resolution information than architectures with downscaling and upsampling. \\ \hline
            Strided convolutions \citep[see][]{dumoulin16}
                & Layer operation &
                The mapping is reliable at reacting to features that do not vary too abruptly in space, i.e.~which are present in several neighboring pixels and can be detected even if the filter center skips some of the pixels. The output is robust towards slight changes of the location of features, and changes of strength/presence of spatially strongly varying features. \\ \hline
            Pooling & Layer operation &
                The output is invariant to slight spatial distortions of the input (slight changes of the location of (deep) features). Features that are sensitive to such distortions can be discarded. \\ \hline
            Stochastic pooling\newline\citep{zeiler13} & Layer operation & 
                The output is robust towards slight changes of the location (like pooling) but also of the strength/presence of (deep) features. \\ \hline
            Training with different kinds of noise (including Dropout; see Section~\ref{sec:data})
                & Noise &
                The mapping is robust to noise: the given class of perturbations of the input or deep features should not affect the output too much. \\ \hline
            Dropout \citep{hinton12,srivastava14}, DropConnect \citep{wan13}, and related methods & Noise & 
                Extracting complementary (non-coadapted) features is helpful. Non-coadapted features are more informative, better disentangle factors of variation. (We want to disentangle factors of variation because they are entangled in different ways in inputs vs.~in outputs.) \newline
                When interpreted as ensemble learning: usual assumptions of ensemble learning (predictions of weak learners have complementary info and can be combined to strong prediction). \\ \hline
            Maxout units\newline\citep{goodfellow13} & Layer operation &
                Assumptions similar to Dropout, with more accurate approximation of model averaging (when interpreted as ensemble learning) \\ \hline
            Skip-connections~\citep{long15,huang16b} & Connections between layers &
                Certain lower-level features can directly be reused in a meaningful way at (several) higher levels of abstraction \\ \hline
            Linearly augmented feed-forward network \citep{vdsmagt12}  & Connections between layers &
                Skip-connections that share weights with the non-skip-connections. Helps against vanishing gradients. Rather changes the learning algorithm than the network mapping. \\ \hline
            Residual learning\newline\citep{he16} & Connections between layers &
                Learning additive difference of a mapping $f$ (or its compositional parts) from the identity mapping is easier than learning $f$ itself. Meaningful deep features can be composed as a sum of lower-level and intermediate-level features. \\ \hline
            Stochastic depth\newline\citep{huang16},\newline DropIn\newline\citep{smith15} & Connections between layers; noise &
                Similar to Dropout: extracting complementary (non-coadapted) features \emph{across different levels of abstraction} is helpful; implicit model ensemble. Similar to Residual learning: meaningful deep features can be composed as a sum of lower-level and intermediate-level features, \emph{with the intermediate-level ones being optional, and leaving them out being meaningful data augmentation}. Similar to Mollifying networks: simplifying random parts of the mapping improves training. \\ \hline
            Mollifying networks\newline\citep{gulcehre16} &
                Connections between layers; noise &                 
                The mapping can be easier approximated by estimating its decreasingly linear simplified version \\ \hline
            Network information criterion \citep{murata94}, Network growing and network pruning \citep[see][Sec.~9.5]{bishop95b} &
                Model selection &
                Optimal generalization is reached by a network that has the right number of units (not too few, not too many) \\ \hline
            Multi-task learning \citep[see][]{caruana98,ruder17} & * & 
                Several tasks can help each other to learn mutually useful feature extractors, as long as the tasks do not compete for resources (network capacity) 
        \\ \hline 
    \end{tabularx}
    } 
    \caption{\label{tbl:arch} Methods based on network architecture, and rough description of assumptions that they encode. There are partial overlaps between some listed methods. For example, Residual learning uses Skip-connections. Many noise-based methods also fit Table~\ref{tbl:data:generic} (cf.~Appendix~\ref{sec:ambi}).}
\end{table}

\paragraph{Assumptions about the mapping}
An~input-output mapping $f_w$ must have certain properties in order to fit the data $P$ well.
Although it may be intractable to enforce the precise properties of an ideal mapping, 
it may be possible to approximate them by simplified assumptions about the mapping.
These properties and assumptions can then be imposed upon model fitting in a hard or soft manner.
This limits the search space of models
and allows finding better solutions.
An~example is the decision about the number of layers and units, which allows the mapping to be neither too simple nor too complex (thus avoiding underfitting and overfitting). 
Another example are certain invariances of the mapping, such as locality and shift-equivariance of feature extraction hardwired in convolutional layers.
Overall, the approach of imposing assumptions about the input-output mapping discussed in this section is the selection of the network architecture~$f$.
The choice of architecture $f$ on the one hand \emph{hardwires} certain properties of the mapping; additionally, in an interplay between $f$ and the optimization algorithm (Section~\ref{sec:opt}), certain weight configurations are more likely accessible by optimization than others, further limiting the likely search space in a \emph{soft} way.
A~complementary way of imposing certain assumptions about the mapping are regularization terms (Section~\ref{sec:regterm}), as well as invariances present in the (augmented) data set (Section~\ref{sec:data}).

Assumptions can be hardwired into the definition of the \emph{operation} performed by certain layers, and/or into the \emph{connections} between layers. This distinction is made in Table~\ref{tbl:arch}, where these and other methods are listed.

In Section~\ref{sec:data} about data, we mentioned regularization methods that transform data in the \hyperref[sec:data:space]{hidden-feature space}.
They can be considered part of the architecture. In other words, they fit both Sections~\ref{sec:data} (data) and~\ref{sec:arch} (architecture). These methods are listed in Table~\ref{tbl:data:generic} with \emph{hidden features} as their \hyperref[sec:data:space]{transformation space}. 

\paragraph{Weight sharing}
Reusing a certain trainable parameter in several parts of the network is referred to as \emph{weight sharing}. This usually makes the model less complex than using separately trainable parameters. An example are convolutional networks~\citep{lecun89}. Here the weight sharing does not merely reduce the number of weights that need to be learned; it also encodes the prior knowledge about the shift-equivariance and locality of feature extraction. Another example is weight sharing in autoencoders.

\paragraph{Activation functions}
Choosing the right activation function is quite important; for example, using Rectified linear units (ReLUs) 
improved the performance of many deep architectures both in the sense of training times and accuracy~\citep{jarrett09,nair10,glorot11}.
The success of ReLUs can be attributed to the fact that they 
help avoiding the vanishing gradient problem,
but also to the fact that they provide more expressive families of mappings (the classical sigmoid nonlinearity can be approximated very well\footnote{Small integrated squared error, small integrated absolute error. A~simple example is~$\mathrm{sigm}(x) \approx \mathrm{ReLU}(x+0.5)-\mathrm{ReLU}(x-0.5)$.} with only two ReLUs, but it takes an infinite number of sigmoid units to approximate a ReLU) and their affine extrapolation to unknown regions of data space seems to provide better generalization in practice than the ``stagnating'' extrapolation of sigmoid units. 
Some activation functions were designed explicitly for regularization. For Dropout, Maxout units~\citep{goodfellow13} allow a more precise approximation of the geometric mean of the model ensemble predictions at test time. Stochastic pooling~\citep{zeiler13}, on the other hand, is a noisy version of max-pooling. The authors claim that this allows modelling distributions of activations instead of taking just the maximum.

\paragraph{Noisy models} Stochastic pooling was one example of a stochastic generalization of a deterministic model. Some models are stochastic by injecting random noise into various parts of the model. The most frequently used noisy model is Dropout~\citep{hinton12, srivastava14}.

\paragraph{Multi-task learning} A special type of regularization is multi-task learning~\citep[see][]{caruana98, ruder17}. It can be combined with semi-supervised learning to utilize unlabeled data on an auxiliary task~\citep{rasmus15}. A similar concept of sharing knowledge between tasks is also utilized in \emph{meta-learning}, where multiple tasks from the same domain are learned sequentially, using previously gained knowledge as bias for new tasks~\citep{baxter00}; and \emph{transfer learning}, where knowledge from one domain is transferred into another domain~\citep{pan10}.

\paragraph{Model selection} The best among several trained models (e.g.~with different architectures) can be selected by evaluating the predictions on a validation set. It should be noted that this holds for selecting the best combination of all techniques (Sections~\mbox{\ref{sec:data}--\ref{sec:opt}}), not just architecture; and that the validation set used for model selection in the ``outer loop'' should be different from the validation set used e.g.~for Early stopping (Section~\ref{sec:opt}), and different from the test set~\citep{cawley10}. However, there are also model selection methods that specifically target the selection of the \emph{number of units} in a specific network architecture, e.g.~using network growing and network pruning~\citep[see][Sec.~9.5]{bishop95b}, or additionally do \emph{not} require a validation set, e.g.~the Network information criterion to compare models based on the training error and second derivatives of the loss function~\citep{murata94}.

\section{Regularization via the error function}
\label{sec:errorterm}
Ideally, the error function $E$ reflects an appropriate notion of quality, and in some cases some assumptions about the data distribution. Typical examples are mean squared error or cross-entropy. The error function $E$ can also have a regularizing effect. An example  is Dice coefficient optimization~\citep{milletari16} which is robust to class imbalance.
Moreover, the overall form of the loss function can be different than Eq.~\eqref{eq:emprisk}. For example, in certain loss functions that are robust to class imbalance, the sum is taken over pairwise combinations $\mathcal{D}\times\mathcal{D}$ of training samples~\citep{yan03}, rather than over training samples. But such alternatives to Eq.~\eqref{eq:emprisk} are rather rare, and similar principles apply.
If additional tasks are added for a regularizing effect (multi-task learning~\citep[see][]{caruana98, ruder17}), then targets $t$ are modified to consist of several tasks, the mapping $f_w$ is modified to produce an according output~$y$, and~$E$~is modified to account for the modified~$t$ and~$y$.
Besides, there are regularization terms that depend on~$\partial E/\partial x$. They depend on $t$ and thus in our definition are considered part of $E$ rather than of $R$, but they are listed in Section~\ref{sec:regterm} among $R$ (rather than here) for a better overview.

\section{Regularization via the regularization term}
\label{sec:regterm}
Regularization can be achieved by adding a regularizer $R$ into the loss function. Unlike the error function $E$ (which expresses consistency of outputs with targets), the regularization term is independent of the targets. Instead, it is used to encode other properties of the desired model, to provide inductive bias (i.e.~assumptions about the mapping other than consistency of outputs with targets).
The value of $R$ can thus be computed for an unlabeled test sample, whereas the value of $E$ cannot.

The independence of $R$ from $t$ has an important implication: it allows additionally using unlabeled samples (semi-supervised learning) to improve the learned model based on its compliance with some desired properties~\citep{sajjadi16}.
For example, semi-supervised learning with ladder networks~\citep{rasmus15} combines a supervised task with an unsupervised auxiliary denoising task in a ``multi-task''
learning fashion. (For alternative interpretations, see Appendix~\ref{sec:ambi}.)
Unlabeled samples are extremely useful when labeled samples are scarce.
A~Bayesian perspective on the combination of labeled and unlabeled data in a semi-supervised manner is offered by~\citet{lassere06}.

A~classical regularizer is \textit{weight decay}~\citep[see][Chap.~7]{plaut86,lang89,dlb}:
\begin{equation}
    \label{eq:wd}
    R(w) = \lambda \frac{1}{2} \lVert w \rVert^2_2 \,,
\end{equation}
where $\lambda$ is a weighting term controlling the importance of the regularization over the consistency. From the Bayesian perspective, weight decay corresponds to using a symmetric multivariate normal distribution as prior for the weights: $p(w) = \mathcal{N}(w|\vec{0},\lambda^{-1}\vec{I})$~\citep{nowlan92}.
Indeed, 
$-\log \mathcal{N}(w|\vec{0},\lambda^{-1}\vec{I}) \propto 
-\log\exp\left(-\frac{\lambda}{2}\lVert w\rVert_2^2\right) = 
\frac{\lambda}{2}\lVert w\rVert_2^2 = R(w)$.
Weight decay has gained big popularity, and it is being successfully used; \citet{krizhevsky12} even observe reduction of the error on the \emph{training} set.

Another common prior assumption that can be expressed via the regularization term is ``smoothness''
of the learned mapping~\citep[see][Section~3.2]{bengio13}: if $x_1 \approx x_2$, then $f_w(x_1) \approx f_w(x_2)$.
It can be expressed by the following loss term:
\begin{equation}
    \label{eq:lossjac}
    R(f_w,x) = \left\lVert J_{f_w}(x) \right\rVert
    _F^2\,,
\end{equation}
where $\lVert \cdot \rVert_F$ denotes the Frobenius norm, and $J_{f_w}(x)$
is the Jacobian of the neural network input-to-output mapping $f_w$ for some fixed network weights $w$.
This term penalizes mappings with large derivatives, and is used in contractive autoencoders~\citep{rifai11a}.

The domain of loss regularizers is very heterogeneous. We propose a natural way to categorize them \textit{by their dependence}. We saw in~Eq.~\eqref{eq:wd} that weight decay depends on $w$ only, whereas the Jacobian penalty in~Eq.~\eqref{eq:lossjac} depends on $w$, $f$, and $x$. More precisely, the Jacobian penalty uses the derivative~$\partial y/\partial x$ of output $y=f_w(x)$
w.r.t.~input~$x$. (We use vector-by-vector derivative notation from matrix calculus, i.e.~$\partial y/\partial x = \partial f_w(x)/\partial x = J_{f_w}$ is the Jacobian of $f_w$ with fixed weights $w$.)
We identify the following dependencies of~$R$:
\begin{itemize}
    \item Dependence on the weights $w$
    \item Dependence on the network output $y=f_w(x)$
    \item Dependence on the derivative $\partial y/\partial w$ of the output $y=f_w(x)$ w.r.t.~the weights~$w$
    \item Dependence on the derivative $\partial y/\partial x$
    of the output $y=f_w(x)$ w.r.t.~the input $x$
    \item Dependence on the derivative $\partial E/\partial x$ of the error term $E$ w.r.t.~the input $x$ ($E$~depends on $t$, and according to our definition such methods belong to Section~\ref{sec:errorterm}, but they are listed here for overview)
\end{itemize}

\renewcommand{\tabularxcolumn}[1]{m{#1}}
\begin{table}
    \setlength\arrayrulewidth{.1pt}
    \scriptsize{
    {
    \begin{tabularx}{\textwidth}{|>{\raggedright\tiny}m{4cm}|>{\tiny}m{4cm}|C{.3cm}|C{0.3cm}|C{0.3cm}|C{0.3cm}|C{0.3cm}|>{\raggedright\arraybackslash\tiny}X|}
        \hline
          &  & \multicolumn{5}{c|}{\scriptsize{Dependency}} &  \\
        \cline{3-7}
        \multirow{-2}{3.5cm}{\scriptsize{Method}} & \multirow{-2}{3cm}{\scriptsize{Description}}& 
            $w$ &
            $y$ &
            $\frac{\partial y}{\partial w}$ &
            $\frac{\partial y}{\partial x}$ &
            $\frac{\partial E}{\partial x}$ &
        \multirow{-2}{*}{\scriptsize{Equivalence}} \\
        \hline \hline
        Weight decay \citep[see][Chap.~7]{plaut86,lang89,dlb} &
            $L^2$ norm on network weights (not biases). Favors smaller weights, thus for usual architectures tends to make the mapping less ``extreme'', more robust to noise in the input. &
            \crss & & & & & \\ \hline
        Weight smoothing \newline \citep{lang89} &
            Penalizes $L^2$ norm of gradients of learned filters, making them smooth. Not beneficial in practice. &
            \crss & & & & & \\ \hline
        Weight elimination \newline \citep{weigend91} &
            Similar to weight decay but favors few stronger connections over many weak ones. &
            \crss & & & & & Goal similar to Narrow and broad Gaussians \\ \hline
        Soft weight-sharing \newline \citep{nowlan92} &
            Mixture-of-Gaussians prior on weights. Generalization of weight decay. Weights are pushed to form a predefined number of groups 
            with similar values. &
            \crss & & & & & \\ \hline
        Narrow and broad Gaussians \citep{nowlan92, blundell15} &
            Weights come from two Gaussians, a narrow and a broad one. Special case of Soft weight-sharing.  &
        \crss & & & & & Goal similar to Weight elimination \\ \hline    
        Fast dropout approximation \newline \citep{wang13} &
            Approximates the loss that dropout minimizes. Weighted $L^2$ weight penalty. Only for shallow networks. &
            \crss & \crss & & & & Dropout \\ \hline
        Mutual exclusivity \newline \citep{sajjadi16} &
            Unlabeled samples push decision boundaries to low-density regions in input space, promoting sharp (confident) predictions. &
            & \crss & & & & \\ \hline
        Segmentation with binary potentials \citep{bentaieb16} &
            Penalty on anatomically implausible image segmentations. &
            & \crss & & & & \\ \hline
        Flat minima search \newline \citep{hochreiter95} &
            Penalty for sharp minima, i.e.\ for weight configurations where small weight perturbation leads to high error increase. Flat minima have low Minimum description length (i.e.~exhibit ideal balance between training error and model complexity) and thus should generalize better~\citep{rissanen86}.
            &
            \crss & & \crss & & & \\ \hline
        Tangent prop \newline \citep{simard92} &
            $L^2$ penalty on directional derivative of mapping in the predefined \emph{tangent directions} that correspond to known input-space transformations. &
            & & & \crss & & Simple data augmentation \\ \hline
        Jacobian penalty
            \newline \citep{rifai11a} &
            $L^2$ penalty on the Jacobian of (parts of) the network mapping---smoothness prior. &
            & & & \crss & & Noise on inputs injection (not exact~\citep[see][]{an96}) \\ \hline
        Manifold tangent classifier \newline \citep{rifai11b} &
            Like tangent prop, but the input ``tangent'' directions are extracted from manifold learned by a stack of contractive autoencoders and then performing SVD of the Jacobian at each input sample. &
            & & & \crss & & \\ \hline
        Hessian penalty \newline \citep{rifai11c} &
            Fast way to approximate $L^2$ penalty of the Hessian of $f$ by penalizing Jacobian with noisy input. &
            & & & \crss & & \\ \hline
        Tikhonov regularizers
            \newline \citep{bishop95} &
            $L^2$ penalty on (up to) $n$-th derivative of the learned mapping
            w.r.t.\ input. &
            & & & \crss & & For penalty on first derivative: noise on inputs injection (not exact~\citep[see][]{an96}) \\ \hline
        Loss-invariant backpropagation \citetext{\citealp[Sec.~3.1]{demyanov15}; \citealp{lyu15}} &
            ($L^2$) norm of gradient of loss w.r.t.\ input. Changes the mapping such that the loss becomes rather invariant to changes of the input. &
            & & & & \crss & Adversarial training \\ \hline
        Prediction-invariant backpropagation
        \newline\citep[Sec.~3.2]{demyanov15}  &
            ($L^2$) norm of directional derivative of mapping w.r.t.\ input in the direction of $x$ causing the largest increase in loss. &
            & & & \crss & \crss & Adversarial training \\ \hline
    \end{tabularx}
    }
    }
    \caption{Regularization terms, with dependencies marked by \crss. Methods that depend on~$\partial E/\partial x$ implicitly depend on targets~$t$ and thus can be considered part of the error function (Section~\ref{sec:errorterm}) rather than regularization term (Section~\ref{sec:regterm}).}
    \label{tbl:regterms}
\end{table}

A~review of existing methods can be found in Table~\ref{tbl:regterms}.
Weight decay seems to be still the most popular of the regularization terms. Some of the methods are equivalent or nearly equivalent to other methods from different taxonomy branches. For example, Tangent prop simulates minimal data augmentation~\citep{simard92}; Injection of small-variance Gaussian noise~\citep{bishop95,an96} is an approximation of Jacobian penalty~\citep{rifai11a}; and Fast dropout~\citep{wang13} is (in shallow networks) a deterministic approximation of Dropout. This is indicated in the \emph{Equivalence} column in Table~\ref{tbl:regterms}.

\section{Regularization via optimization}
\label{sec:opt}
The last class of the regularization methods according to our taxonomy is the regularization through optimization. Stochastic gradient descent (SGD)~\citep[see][]{bottou98} (along with its derivations) is the most frequently used optimization algorithm in the context of deep neural networks and is the center of our attention. We also list some alternative methods below.

Stochastic gradient descent is an iterative optimization algorithm using the following update rule:
\begin{equation}
\label{eq:sgd}
w_{t+1} = w_t - \eta_t \nabla_w \L(w_t, d_t) ,
\end{equation}
where $\nabla \L(w_t, d_t)$
is the gradient of the loss
$\mathcal{L}$ 
evaluated on a mini-batch $d_t$ from the training set $\mathcal{D}$. It is frequently used in combination with \textit{momentum} and other tweaks improving the convergence speed~\citep[see][]{wilson17}.
Moreover, the noise induced by the varying mini-batches helps the algorithm escape saddle points~\citep{ge15}; this can be further reinforced by adding supplementary gradient noise~\citep{neelakantan15,chaudhari15}.

If the algorithm reaches a low training error in a reasonable time (linear in the size of the training set, allowing multiple passes through $\mathcal{D}$), the solution generalizes well under certain mild assumptions; in that sense SGD works as an \textit{implicit regularizer}: a~short training time prevents overfitting even without any additional regularizer used~\citep{hardt16}.
This is in line with~\citep{zhang16} who find in a series of experiments that regularization (such as Dropout, data augmentation, and weight decay)
is by itself neither necessary nor sufficient for good generalization.

We divide the methods into three groups: initialization/warm-start methods, update methods, and termination methods, discussed in the following.

\paragraph{Initialization and warm-start methods}
\label{sec:opt:init}
These methods affect the initial selection of the model weights.
Currently the most frequently used method is sampling the initial weights from a carefully tuned distribution. There are multiple strategies based on the architecture choice, aiming at keeping the variance of activations in all layers around~$1$, thus preventing vanishing or exploding activations (and gradients) in deeper layers~\citetext{\citealp[][Sec.~4.2]{glorot10}; \citealp{he15}}.

Another (complementary) option is \emph{pre-training}
on different data, or with a different objective, or with partially different architecture. This can prime the learning algorithm towards a good solution before the fine-tuning on the actual objective starts. Pre-training the model on a different task in the same domain may lead to learning useful features, making the primary task easier.
However, pre-trained models are also often misused as a lazy approach to problems where training from scratch or using thorough domain adaptation, transfer learning, or multi-task learning methods 
would
be worth trying. On the other hand, pre-training or similar techniques may be a useful \emph{part} of such methods.

Finally, with some methods such as Curriculum learning~\citep{bengio09}, the transition between pre-training and fine-tuning is smooth. We refer to them as \emph{warm-start methods}.

\begin{itemize}
    \item Initialization without pre-training
    \begin{itemize}
        \item Random weight initialization~\citetext{\citealp[][p.~330]{rumelhart86}; \citealp{glorot10}; \citealp{he15}; \citealp{hendrycks16}}
        \item Orthogonal weight matrices~\citep{saxe13}
        \item Data-dependent weight initialization~\citep{kraehenbuehl15}
    \end{itemize}
    \item Initialization with pre-training
    \begin{itemize}
        \item Greedy layer-wise pre-training~\citep{hinton06, bengio07, erhan10} (has become less important due to advances (e.g.~ReLUs) in effective end-to-end training that optimizes all parameters simultaneously)
        \item Curriculum learning~\citep{bengio09}
        \item Spatial contrasting~\citep{hoffer16}
        \item Subtask splitting~\citep{gulcehre13}
    \end{itemize}
\end{itemize}

\paragraph{Update methods}
\label{sec:opt:update}
This class of methods affects individual weight updates. There are two complementary subgroups: \textit{Update rules} modify the form of the update formula; \textit{Weight and gradient filters} are methods that affect the value of the gradient or weights, which are used in the update formula, e.g.~by injecting noise into the gradient~\citep{neelakantan15}.

Again, it is not entirely clear which of the methods only speed up the optimization and which actually help the generalization. \citet{wilson17} show that some of the methods such as AdaGrad or Adam even lose the regularization abilities of SGD.

\begin{SCfigure}
    \centering
    \includegraphics[width=5cm]{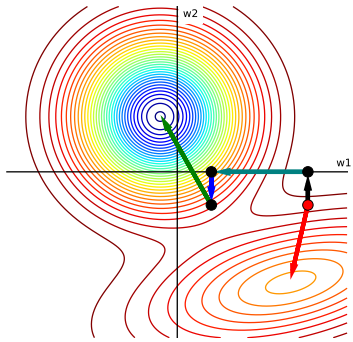}
    \caption{Effect of Dropout on weight optimization. Starting from the current weight configuration (red dot), all weights of certain neurons are set to zero (black arrow), descent step is performed in that subspace (teal arrow), and then the discarded weight-space coordinates are restored (blue arrow).}
    \label{fig:dropout}
\end{SCfigure}

\begin{itemize}
    \item Update rules
    \begin{itemize}
        \item Momentum, Nesterov's accelerated gradient method, AdaGrad, AdaDelta, RMS\-Prop, Adam---overview in~\citep{wilson17}
        \item Learning rate schedules~\citep{girosi95,hoffer17}
        \item Online batch selection~\citep{loshchilov15}
        \item SGD alternatives: L-BFGS~\citep{liu89,le11}, Hessian-free methods~\citep{martens10}, Sum-of-functions optimizer~\citep{sohl14}, ProxProp~\citep{frerix17}
    \end{itemize}
    \item Gradient and weight filters
    \begin{itemize}
        \item Annealed Langevin noise~\citep{neelakantan15}
        \item AnnealSGD~\citep{chaudhari15}
        \item Dropout~\citep{hinton12, srivastava14} corresponds to optimization steps in subspaces of weight space, see Figure~\ref{fig:dropout}
        \item Annealed noise on targets~\citep{wang99} (works as noise on gradient, but belongs rather to data-based methods, Section~\ref{sec:data})
    \end{itemize}
\end{itemize}

\paragraph{Termination methods}
\label{sec:opt:termination}

There are numerous possible stopping criteria and selecting the right moment to stop the optimization procedure may improve the generalization by reducing the error caused by the discrepancy between the minimizers of expected and empirical risk: The network first learns general concepts that work for all samples from the ground truth distribution $P$ before fitting the specific sample $\mathcal{D}$ and its noise~\citep{krueger17}.

The most successful and popular termination methods put a portion of the labeled data aside as a \emph{validation set} and use it to evaluate performance (\textit{validation error}). The most prominent example is Early stopping~\citep[see][]{prechelt98}. In scenarios where the training data are scarce it is possible to resort to termination methods that do \emph{not} use a validation set. The simplest case is fixing the number of passes through the training set.

\begin{itemize}
    \item Termination using a validation set
    \begin{itemize}
        \item Early stopping~\citep[see][]{morgan90,prechelt98}
        \item Choice of validation set size based on test set size~\citep{amari97}
    \end{itemize}
    \item Termination \emph{without} using a validation set
    \begin{itemize}
        \item Fixed number of iterations
        \item Optimized approximation algorithm~\citep{liu08}
    \end{itemize}
\end{itemize}

\section{Recommendations, discussion, conclusions}
\label{sec:discussion}

We see the main benefits of our taxonomy to be two-fold: Firstly, it provides an overview of the existing techniques to the users of regularization methods and gives them a better idea of how to choose the ideal combination of regularization techniques for their problem. Secondly, it is useful for development of new methods, as it gives a comprehensive overview of the main principles that can be exploited to regularize the models. We summarize our recommendations in the following paragraphs:

\paragraph{Recommendations for users of existing regularization methods} 
Overall, using the information contained in data as well as prior knowledge as much as possible, and primarily starting with popular methods, the following procedure can be helpful:

\begin{itemize}
    \item Common recommendations for the first steps:
    \begin{itemize}
        \item Deep learning is about disentangling the factors of variation. An appropriate data representation should be chosen;
        \emph{known} meaningful data transformations should \emph{not} be outsourced to the learning. Redundantly providing the same information in several representations is okay.
        \item Output nonlinearity and error function should reflect the learning goals.
        \item A good starting point are techniques that usually work well (e.g.~ReLU, successful architectures). 
        Hyperparameters (and architecture) can be tuned jointly, but ``lazily'' (interpolating/extrapolating from experience instead of trying too many combinations).
        \item Often it is helpful to start with a simplified dataset (e.g.~fewer and/or easier samples) and a simple network, and after obtaining promising results gradually increasing the complexity of both data and network while tuning hyperparameters and trying regularization methods.
    \end{itemize}
    \item Regularization via data:
    \begin{itemize}
        \item When not working with nearly infinite/abundant data:
        \begin{itemize}
            \item Gathering more real data (and using methods that take its properties into account) is advisable if possible:
            \begin{itemize}
                \item Labeled samples are best, but unlabeled ones can also be helpful (compatible with semi-supervised learning).
                \item Samples from the same domain are best, but samples from similar domains can also be helpful (compatible with domain adaptation and transfer learning).
                \item Reliable high-quality samples are best, but lower-quality ones can also be helpful (their confidence/importance can be adjusted accordingly).
                \item Labels for an additional task can be helpful (compatible with multi-task learning).
                \item Additional input features (from additional information sources) and/or data preprocessing (\hyperref[sec:data:universality]{i.e.~domain-specific data transformations}) can be helpful (the network architecture needs to be adjusted accordingly).
            \end{itemize}
            \item Data augmentation  (e.g.~target-preserving handcrafted domain-specific transformations) can well
            compensate for limited data. If natural ways to augment data (to mimic natural transformations sufficiently well) are known, they can be tried (and combined).
            \item If natural ways to augment data are unknown or turn out to be insufficient,
            it may be possible to infer the transformation from data (e.g.~learning image-deformation fields) if a sufficient amount of data is available for that.
        \end{itemize}  
        \item Popular \hyperref[sec:data:universality]{generic} methods (e.g.~advanced variants of Dropout) often also help.
    \end{itemize}
    \item Architecture and regularization terms:
    \begin{itemize}
        \item Knowledge about possible meaningful properties of the mapping can be used to e.g.~hardwire invariances (to certain transformations) into the architecture, or be formulated as regularization terms.
        \item Popular methods may help as well (see Tables~\ref{tbl:arch}--\ref{tbl:regterms}), but should be chosen to match the assumptions about the mapping (e.g.~convolutional layers are fully appropriate only if local and shift-equivariant feature extraction on regular-grid data
        is desired).
    \end{itemize}
    \item Optimization:
    \begin{itemize}
        \item Initialization: Even though pre-trained ready-made models greatly speed up prototyping, training from a good random initialization should also be considered.
        \item Optimizers: Trying a few different ones, including advanced ones (e.g.~Nesterov momentum, Adam, ProxProp), may lead to improved results. Correctly chosen parameters, such as learning rate, usually make a big difference.
    \end{itemize}
\end{itemize}

\paragraph{Recommendations for developers of novel regularization methods} Getting an overview and understanding the reasons for the success of the best methods is a great foundation.
Promising empty niches (certain combinations of taxonomy properties) exist that can be addressed.
The assumptions to be imposed upon the model can have a strong impact on most elements of the taxonomy. Data augmentation is more expressive than loss terms (loss terms enforce properties only in infinitesimally small neighborhood of the training samples; data augmentation can use rich transformation parameter distributions). 
Data and loss terms impose assumptions and invariances in a rather soft manner, and their influence can be tuned, whereas hardwiring the network architecture is a harsher way to impose assumptions. Different assumptions and options to impose them have different advantages and disadvantages.

\paragraph{Future directions for data-based methods} There are several promising directions that in our opinion require more investigation: \hyperref[sec:data:sampling]{Adaptive sampling of~$\theta$} might lead to lower errors and shorter training times~\citep{fawzi16}
(in~turn, shorter training times may additionally work as implicit regularization~\citep{hardt16}, see also Section~\ref{sec:opt}). Secondly, learning \hyperref[sec:data:cond]{class-dependent transformations} (i.e.~$p(\theta|t)$)
in our opinion might lead to more plausible samples. Furthermore, the field of adversarial examples (and network robustness to them) is gaining increased attention after the recently sparked discussion on real-world adversarial examples and their robustness/invariance to transformations such as the change of camera position~\citep{lu17,athalye17}.
Countering strong adversarial examples may require better regularization techniques.

\paragraph{Summary}
In this work we proposed a broad definition of regularization for deep learning, identified five main elements of neural network training (data, architecture, error term, regularization term, optimization procedure), described regularization via each of them, including a further, finer taxonomy for each, and presented example methods from these subcategories. Instead of attempting to explain referenced works in detail, we merely pinpointed their properties relevant to our categorization.
Our work
demonstrates some links between existing methods. Moreover, our systematic approach enables the discovery of new, improved regularization methods by combining the best properties of the existing ones.

\section*{Acknowledegments}
We thank Antonij Golkov for valuable discussions. Grant support: ERC Consolidator Grant ``3DReloaded''.

\bibliographystyle{apalike}
\bibliography{bibliography}

\appendix
\scriptsize
\section{Ambiguities in the taxonomy}
\label{sec:ambi}

Although our proposed taxonomy seems intuitive, there are some ambiguities: Certain methods have multiple interpretations matching various categories. Viewed from the exterior, a neural network maps inputs~$x$ to outputs~$y$.
We formulate this as $y=f_w(\tau_\theta(x))$ for \hyperref[sec:data:space]{transformations $\tau_\theta$ in input space} (and similarly for \hyperref[sec:data:space]{hidden-feature space}, where $\tau_\theta$ is applied in between layers of the network $f_w$). However, how to split this $x$-to-$y$ mapping into ``the $\tau_\theta$ part'' and ``the $f_w$ part'',
and thus into Section~\ref{sec:data} vs.~Section~\ref{sec:arch}, is ambiguous and up to one's taste and goals.
In our choices (marked with ``\Checkedbox'' below), we attempt to use common notions
and Occam's razor.

\begin{itemize}
    \item Ambiguity of attributing noise to~$f$, or to~$w$, or to data transformations $\tau_\theta$:
    \begin{itemize}
        \item Stochastic methods such as
        Stochastic depth~\citep{huang16} can have several interpretations if stochastic transformations are allowed for~$f$ or~$w$:
        \begin{itemize}
            \item[\Checkedbox] Stochastic transformation of the architecture~$f$ (randomly dropping some connections), Table~\ref{tbl:arch}
            \item[\HollowBox] Stochastic transformation of the weights~$w$ (setting some weights to~$0$ in a certain random pattern)
            \item[\HollowBox] Stochastic transformation $\tau_\theta$ of data in \hyperref[sec:data:space]{hidden-feature space}; \hyperref[sec:data:cond]{dependence is~$p(\theta)$}, described in Table~\ref{tbl:data:generic} for completeness
        \end{itemize}
    \end{itemize}
    \item Ambiguity of splitting~$\tau_\theta$ into~$\tau$ and~$\theta$:
    \begin{itemize}
        \item Dropout:
        \begin{itemize}
            \item[\Checkedbox] Parameters $\theta$ are the dropout mask; \hyperref[sec:data:cond]{dependence is $p(\theta)$}; transformation~$\tau$ applies the dropout mask to the hidden features
            \item[\HollowBox] Parameters $\theta$ are the seed state of a pseudorandom number generator; \hyperref[sec:data:cond]{dependence is $p(\theta)$}; transformation~$\tau$ internally generates the random dropout mask from the random seed and applies it to the hidden features
        \end{itemize}
        \item Projecting dropout noise into input space~\citep[Sec.~3]{bouthillier15} can fit our taxonomy in different ways by defining $\tau$ and $\theta$ accordingly. It can have similar interpretations as Dropout above (if $\tau$ is generalized to allow for dependence on $x,f,w$), but we prefer the third interpretation without such generalizations:
        \begin{itemize}
            \item[\HollowBox] Parameters $\theta$ are the dropout mask (to be applied in a hidden layer); \hyperref[sec:data:cond]{dependence is $p(\theta)$}; transformation~$\tau$ transforms the input to mimic the effect of the mask
            \item[\HollowBox] Parameters $\theta$ are the seed state of a pseudorandom number generator; \hyperref[sec:data:cond]{dependence is $p(\theta)$}; transformation~$\tau$ internally generates the random dropout mask from the random seed and transforms the input to mimic the effect of the mask
            \item[\Checkedbox] Parameters $\theta$ describe the transformation of the input in any formulation; \hyperref[sec:data:cond]{dependence is $p(\theta|x,f,w)$}; transformation~$\tau$ merely applies the transformation in input space
        \end{itemize}
    \end{itemize}
    \item Ambiguity of splitting the network operation $f_w$ into layers: There are several possibilities to represent a function (neural network) as a composition (or directed acyclic graph) of functions (layers).
    \item
    Many of the input and hidden-feature transformations (Section~\ref{sec:data}) can be considered layers of the network (Section~\ref{sec:arch}). In fact, the term ``layer'' is not uncommon for Dropout or Batch normalization.
    \item The usage of a trainable parameter in \emph{several} parts of the network is called weight sharing. However, some mappings can be expressed with two equivalent formulas such that a parameter appears only once in one formulation, and several times in the other.
    \item Ambiguity of $E$ vs.~$R$: Auxiliary denoising task in ladder networks~\citep{rasmus15} and similar autoencoder-style loss terms can be interpreted in different ways:
    \begin{itemize}
        \item[\Checkedbox] Regularization term $R$ without given auxiliary targets $t$
        \item[\HollowBox] The ideal reconstructions can be considered as targets $t$ (if the definition of ``targets'' is slightly modified) and thus the denoising task becomes part of the error term $E$
    \end{itemize}
\end{itemize}

\section{Data-augmented loss function}
\label{sec:aug}
To understand the success
of target-preserving data augmentation methods, we consider the \textit{data-augmented loss function}, which we obtain by replacing the training samples $(x_i,t_i)\in\mathcal{D}$ in the empirical risk loss function (Eq.~\eqref{eq:emprisk}) by augmented training samples $(\tau_\theta(x_i), t_i)$:
\begin{equation}
\label{eq:augloss}
\begin{split}
\hat\L_A &= \frac{1}{|\mathcal{D}|} \sum_{(x_i,t_i)\in \mathcal{D}} \E_{\theta} \Big[  \ell\big(\tau_\theta(x_i), t_i\big) \Big] \\
         &= \frac{1}{|\mathcal{D}|} \sum_{(x_i,t_i)\in \mathcal{D}} \int\limits_\Theta \Big( \ell\big(\tau_\theta(x_i), t_i\big) \Big) p(\theta) \dif \theta,
\end{split}
\end{equation}

where we have replaced the inner part ($E$~and~$R$) of the loss function by $\ell$ to simplify the notation. Moreover,~$\hat \L_A$ can be rewritten as
\begin{equation}
\label{eq:aug2}
\begin{split}
\hat\L_A &= \iint\limits_{X,T} \frac{1}{|\mathcal{D}|} \sum_{(x_i,t_i)\in \mathcal{D}} \int\limits_\Theta \ell(x,t) \; p(\theta) \; \delta\big(x - \tau_\theta (x_i)\big) \; \delta(t-t_i) \dif \theta \dif t \dif x \\
         &= \iint\limits_{X,T} \ell(x,t) \Bigg[ \frac{1}{|\mathcal{D}|} \sum_{(x_i,t_i)\in \mathcal{D}} \int\limits_\Theta \delta\big(x-\tau_\theta (x_i)\big) \; \delta(t-t_i) \; p(\theta) \dif \theta \Bigg] \dif t \dif x \\
         &= \iint\limits_{X,T} \ell(x,t) \; q(x,t) \dif t \dif x ,
\end{split}
\end{equation}
where $\delta(x)$ is the Dirac delta function: $\delta(x) = 0\,\forall\,x \ne 0$ and $\int\delta(x)\dif x=1$; 
and $q(x,t)$ is defined as
\begin{equation}
\label{eq:augdensity}
q(x,t) = \frac{1}{|\mathcal{D}|} \sum_{(x_i,t_i)\in \mathcal{D}} \; \int\limits_\Theta \delta\big(x-\tau_\theta (x_i)\big) \; \delta(t-t_i) \; p(\theta) \dif \theta .
\end{equation}
Since $q$ is non-negative and $\iint q(x,t) \dif x \dif t = 1$, it is a valid probability density function
inducing the distribution~$Q$ of augmented data. Therefore,
\begin{equation}
\label{eq:aug4}
\hat\L_A = \E_{(x,t)\sim Q} \big[ \ell(x,t) \big] .
\end{equation}
When $Q=P$, Eq.~\eqref{eq:aug4} becomes the expected risk~\eqref{eq:exprisk}. We can show how this is related to \textit{importance sampling}:
\begin{equation}
\label{eq:importance}
\begin{split}
\L &= \E_{(x,t)\sim P} \big[ \ell(x,t) \big]\\
   &= \iint\limits_{X,T} \ell(x,t) p(x,t) \dif t \dif x \\
   &= \iint\limits_{X,T} \ell(x,t) \frac{p(x,t)}{q(x,t)} q(x,t) \dif t \dif x\\
   &= \E_{(x,t)\sim Q} \bigg[ \ell(x,t) \frac{p(x,t)}{q(x,t)} \bigg] \\
   &\ne \E_{(x,t)\sim Q} \big[ \ell(x,t) \big] \\
   &= \hat\L_A   .  
\end{split}
\end{equation}
The difference between $\L$ and $\hat\L_A$ is the re-weighting term $p(x,t)/q(x,t)$ identical to the one known from importance sampling~\citep[see][]{bishop95b}. The more similar $Q$ is to $P$ (i.e.~the closer $Q$ models the ground truth distribution $P$), the more similar the augmented-data loss $\hat\L_A$ is to the expected loss $\L$. We see that data augmentation tries to simulate the real distribution~$P$ by creating new samples from the training set~$\mathcal{D}$, bridging the gap between the expected and the empirical risk.

\end{document}